%% file: main.tex
\documentclass[10pt,twocolumn,letterpaper]{article}

\usepackage[pagenumbers]{cvpr} 

\input{preamble}
\definecolor{cvprblue}{rgb}{0.21,0.49,0.74}
\usepackage[pagebackref,breaklinks,colorlinks,allcolors=cvprblue]{hyperref}

\def\paperID{16950} 
\def\confName{CVPR}
\def\confYear{2026}

\title{Text-Only Training for Image Captioning with Retrieval Augmentation and Modality Gap Correction}

\author{Rui Fonseca \quad Bruno Martins \quad Gil Rocha\\
INESC-ID, Instituto Superior Tecnico, University of Lisbon\\
{rui.f.fonseca@tecnico.ulisboa.pt}
}

\begin{document}
\maketitle
\input{sec/0_abstract}    
\input{sec/1_introduction}
\input{sec/2_relatedwork}
\input{sec/3_proposedmethod}
\input{sec/4_evaluation}
\input{sec/5_conclusion}
{
    \small
    \bibliographystyle{ieeenat_fullname}
    \bibliography{main}
}

\clearpage
\newpage
\input{sec/appendix}


\end{document}

%% file: sec/0_abstract.tex
\begin{abstract}
Image captioning has drawn considerable attention from the natural language processing and computer vision fields. Aiming to reduce the reliance on curated data, several studies have explored image captioning without any humanly-annotated image-text pairs for training, although existing methods are still outperformed by fully supervised approaches. This paper proposes TOMCap, i.e., an improved text-only training method that performs captioning without the need for aligned image-caption pairs. The method is based on prompting a pre-trained language model decoder with information derived from a CLIP representation, after undergoing a process to reduce the modality gap. We specifically tested the combined use of retrieved examples of captions, and latent vector representations, to guide the generation process. Through extensive experiments, we show that TOMCap outperforms other training-free and text-only methods. We also analyze the impact of different choices regarding the configuration of the retrieval-augmentation and modality gap reduction components.
\end{abstract}

%% file: sec/1_introduction.tex
\section{Introduction}
\label{sec:intro}

Image captioning concerns generating textual descriptions for input images. Due to the vast amount of applications, the task has gathered a substantial amount of attention from the natural language processing and computer vision communities. Models that rely on encoder-decoder frameworks have significantly advanced the state-of-the-art~\cite{alayrac2022flamingo,li2023blip}. However, these models are usually trained with full supervision and require large-scale humanly-annotated training data (i.e., manually annotated image-captions pairs).

To facilitate model development, several studies have proposed encoder–decoder architectures that re-use off-the-shelf pre-trained vision encoders, such as CLIP~\cite{radford2021learning}, and language decoders, such as GPT2~\cite{radford2019language}. In these approaches, the parameters of the pre-trained models are kept frozen, and only the mapping between the two components is trained. Some authors have further proposed the use of retrieval-augmented generation to enhance efficiency while enabling domain transfer and exploration of captioning data in a training-free manner~\cite{ramos2023smallcap,ramos2024mpaella}. 

To address data limitations and improve generalization to different domains, other studies have explored image captioning without relying on any human-annotated image–text pairs. These models can be broadly categorized into two groups, namely training-free and text-only methods. Training-free approaches achieve zero-shot captioning by leveraging pre-trained models without additional fine-tuning (e.g., a pre-trained vision–language model such as CLIP is used to guide a pre-trained language model like GPT2 to generate a caption corresponding to the given image). Conversely, text-only methods fine-tune some of the model components using high-quality text corpora, without requiring the corresponding images during training. 

Despite significant progress, training-free methods remain prone to hallucination issues, while text-only approaches, although capable of achieving strong results, are still outperformed by fully supervised models. A significant issue that limits the performance of text-only training methods is the well known CLIP modality gap~\cite{liang2022mind}, in which the embeddings corresponding to images and texts do not perfectly align in the same vector space.

This paper proposes Text-Only Training Captioning with Modality Gap Correction (TOMCap), i.e. an enhanced text-only training framework that unifies ideas from prior research, including retrieval-augmented generation as in SmallCap~\cite{ramos2023smallcap} and LMCap~\cite{ramos2023lmcap}, the decoding of latent representations from CLIP encoders~\cite{gu2022can,nukrai2022text,qiu2024mining,wang2024text}, and the application of correction mechanisms to CLIP embeddings to minimize the modality gap~\cite{MixedModalitySearch}. Instead of relying solely on a corpus of textual captions to train the decoder, our method additionally utilizes the text-only corpus as a retrieval datastore to adjust the decoder using retrieved captions as targets. We also incorporate Low-Rank Adaptation (LoRA)~\cite{hu2021lora} to fine-tune the model without retraining all parameters The decoder is thus optimized to generate captions conditioned on (a) semantically similar captions obtained via retrieval, and (b) latent representations of the input. Both conditioning elements are derived from CLIP embeddings, either from textual captions during training or from image features during inference, while multiple strategies are applied to mitigate the CLIP modality gap, as detailed in Section~\ref{sec:clipfix}. Finally, combining retrieval augmentation with LoRA-based fine-tuning helps preserve the linguistic knowledge of the pretrained language model, while improving its visual grounding capability.

Experiments with MSCOCO~\cite{lin2014microsoft} and NoCaps~\cite{agrawal2019nocaps} show that TOMCap outperforms previous training-free and text-only training methods, particularly in terms of generalization in NoCaps. The retrieval of similar captions is the most impactful aspect in the proposed combination, and we also analyzed the impact of different choices regarding the configuration of the retrieval-augmentation component, including the number of retrieval examples and trade-offs between caption similarity and diversity.

%% file: sec/2_relatedwork.tex
\section{Related Work}
\label{sec:relatedwork}

This section covers relevant previous work related to the proposed approach. 

\subsection{Training-Free Captioning}

Previous studies have shown that captions for a given image can be generated by re-purposing text-image encoders, such as CLIP~\cite{radford2021learning}, to guide a decoder, such as GPT2~\cite{radford2019language}. This approach has the advantage of not requiring any task-specific training. An example of one such architecture is ZeroCap~\cite{tewel2022zerocap}, which modified a GPT2 decoder is used to perform the task of captioning, without any model re-training or fine-tuning. This was achieved via the addition of a context cache, which is updated with the guidance from CLIP and GPT2 for every prediction.

Several other previous strategies also do not directly involve the use of image data, relying instead on textual information alone. For instance~\citet{zeng2022socratic} proposed the Socratic models framework, where different pre-trained models can communicate via zero-shot prompting, without any multimodal training. For instance, for the task of image captioning, a GPT language model can be prompted with information about the input image (e.g., information about the number of people present in the image, places, objects, and general classes associated with the image), as obtained with a pre-trained CLIP model. In turn, Ramos et al.~\cite{ramos2023lmcap} proposed LMCap, building on similar ideas to those of Socratic models. LMCap is an image-blind method that generates captions only with basis on textual information provided as input. In this case, CLIP is first used to retrieve captions from similar images, and these captions are then combined into a prompt for a GPT2 language model decoder. 

\subsection{Text-Only Training for Captioning}

Rather than using a zero-shot approach, text-only training might be more suitable for the task of image captioning~\cite{nukrai2022text}. This means that a decoder can be trained to generate captions conditioned on the outputs of a frozen CLIP text encoder, this way only using textual data ~\cite{gu2022can,nukrai2022text,ifcap2024}. The training becomes the reconstruction of the textual data, which is performed based on the embeddings generated by a CLIP encoder. At inference time, the model substitutes the textual embedding with an image embedding obtained via the corresponding CLIP image encoder. This allows the model to, in theory, train only on texts and then switch to images during inference. However, the presence of a modality gap between text and image representations is a critical issue that affects the performance of these models. To address this, most text-only training approaches use noise injections during training to bridge the gap (e.g., through the introduction of Gaussian noise into the embeddings generated by CLIP during training)~\cite{gu2023language}.

More recent text-only training approaches use the available textual training data to generate synthetic images. These approaches provide a pseudo multimodal training setup without requiring large-scale paired data, leveraging text-to-image diffusion models or pretrained generative priors to synthesize visual representations aligned with the textual content, effectively serving as a bridge between purely textual training and multimodal understanding~\cite{jiang-etal-2023-exploiting, liu2024improving}.

\subsection{Retrieval-Augmented Captioning}

To enhance caption generation procedures, several authors have explored the incorporation of additional contextual information about the image into a prompt, typically achieved through the use of an external datastore. These models leverage retrieval mechanisms to augment and guide the caption generation process~\cite{sarto2022retrieval,ramos2023retrieval,ramos2023smallcap,ramos2023lmcap,ramos2024mpaella}.

A representative example is SmallCap~\cite{ramos2023smallcap}, which employs a frozen CLIP encoder to compare caption embeddings against examples of captions within a datastore, enabling the retrieval of semantically similar captions that are subsequently incorporated into a prompt for a GPT2 decoder. Moreover, SmallCap leverages the latent representations of CLIP embeddings to guide caption generation by introducing cross-attention layers in the decoder, facilitating a mapping between the visual and textual modalities.

While retrieval-augmented captioning has demonstrated promising results, it remains sensitive to several factors. The modality gap between visual and textual embeddings, as well as potential limitations in the coverage or domain alignment of the datastore, can significantly affect performance. Prior work has shown that aspects such as the semantic similarity of retrieved captions, and even the order in which they are incorporated into the prompt, can influence the quality of the generated captions~\cite{li-etal-2024-understanding-retrieval}.

%% file: sec/3_proposedmethod.tex
\section{Proposed Method}
\label{sec:proposedmethod}

\begin{figure*}[t!]
  \centering
  \includegraphics[width=\linewidth]{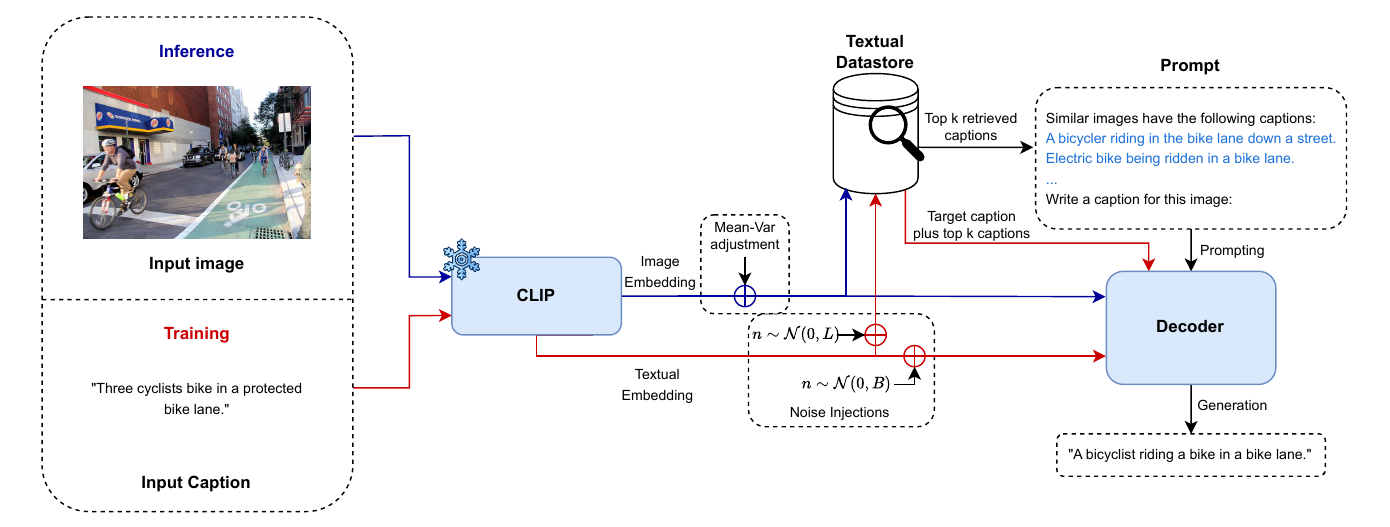}
  \caption{Overview of the training (red) and inference (blue) procedures associated to TOMCap.}
  \label{fig:base_model}
\end{figure*}

The approach presented here, which we name Text-Only Training Captioning with Modality Gap Correction (TOMCap), combines retrieval-augmented generation~\cite{ramos2023smallcap,ramos2023lmcap}, the decoding of latent representations from CLIP, and a correction mapping applied to CLIP embeddings~\cite{MixedModalitySearch}. The proposed method relies on textual information during training, while, at inference time, it uses a textual prompt together with visual information encoded by CLIP.

First, prior to training, examples of image captions are encoded using the CLIP text encoder, producing vector representations. To bridge the modality gap, a correction function is applied to these vectors, followed by the injection of Gaussian noise. The resulting representations are stored in a datastore and used later to retrieve semantically similar captions. The retrieved captions and the CLIP embeddings are then used as inputs to a language model decoder. This decoder, in addition to processing the prompt, is extended with cross-attention layers that take as key and value vectors the same corrected embeddings used during the retrieval step. The parameters of both CLIP and the language model decoder remain frozen throughout training, while only the newly added cross-attention mechanism and Low-Rank Adaptation (LoRA) layers~\cite{hu2021lora} are updated. By employing LoRA instead of full-parameter fine-tuning, the method reduces computational costs and helps retain the general linguistic knowledge of the pretrained language model, thereby mitigating catastrophic forgetting.

During inference, given an input image, the model encodes the image using CLIP, producing a vector representation that is then corrected to align with the textual embedding space. This vector, similarly to the representations used during training, is employed to retrieve semantically similar captions from the pre-encoded datastore. These captions are then used to construct a prompt and as input to the cross-attention mechanism in the decoder. The combination of retrieved captions within the prompt also conditions the caption generation process. The main components of our approach are illustrated in Figure~\ref{fig:base_model}, and further details are provided in the following sections.

\subsection{Building CLIP Representations}
\label{sec:clipfix}

To encode the images (during inference) and the captions (during training), we use a SigLIP2 L/16\footnote{\tiny \url{https://huggingface.co/google/siglip2-large-patch16-512}} model whose parameters are kept frozen. This model generates embeddings of dimentionality 1024, which are then adjusted to minimize the CLIP modality gap.

Previous studies have shown that a significant modality gap exists between the vector representations generated by the vision-language CLIP encoders~\cite{gu2023language}. Notably, these studies show that for a given image-caption pair, the corresponding embeddings lie within a small ball of radius $\epsilon$. A method previously proposed to reduce $\epsilon$ is to subtract the mean values of each modality from the pair~\cite{MixedModalitySearch}. This effectively aligns the mean vector of the images with that of the texts. We further develop this idea by also considering the standard deviation, as ignoring it can result in mismatched ranges for the image and text embeddings. 

Formally, let $I_n$ and $T_n$ denote an image and its corresponding caption. Applying the CLIP encoder produces embeddings $e_d^{I_n}$ and $e_d^{T_n}$ for the image and caption, respectively, where $d$ is the embedding dimensionality. Given independent sets of images and captions, we can construct two distributions, $E^I$ and $E^T$, derived from the embeddings of each image and caption within the sets. From these distributions, we compute the mean and standard deviation for each dimension in the embedding space: $\mu_d^{I}$, $\mu_d^{T}$, $\sigma_d^{I}$, and $\sigma_d^{T}$, where $d$ again consists of the embedding dimensionality. Assuming independence between each embedding dimension, we can correct the text embeddings $e_d^{T_n}$ using the following formula:
\begin{equation}
    e^{T'_n}_{d} = (e^{T_n}_d - \mu_d^{T}) \times \frac{\sigma_d^{I}}{\sigma_d^{T}} + \mu_d^{I},
\end{equation}
which places the corrected text embedding $e^{T'_n}_{d}$ closer to the corresponding image embedding $e^{I_n}_d$, thus reducing the modality gap to a smaller radius.

After applying this correction, Gaussian noise is also injected into each dimension to further reduce the remaining modality gap, as suggested by Nukrai et al.~\cite{nukrai2022text}. The noise is introduced by multiplying a standard normal distribution by a vector of standard deviations for each dimension. In our experiments, the standard deviations are either fixed or randomly generated from a normal distribution scaled by a value $B$ or $L$ when applying the noise to the cross-attention inputs or retrieval results, respectively. The addition of two independent noise injections allows for different amounts of noise to be introduced in either component, to take into account the sensitivity to noise of each component.

\subsection{Retrieval of Similar Captions}

The SigLIP2~\cite{tschannen2025siglip2multilingualvisionlanguage} model used to encode captions and images is also employed for text-to-text and image-to-text retrieval during the training and inference stages. In addition to encoding the input images or captions, we also use this CLIP model to encode the instances within a datastore composed of a diverse set of captions, which are then used for retrieval and for constructing the prompt. The datastore is indexed offline, following previous work such as LMCap~\cite{ramos2023lmcap}. We use FAISS nearest neighbor search, specifically with  {\tt IndexFlatL2} for a datastore containing approximately 16M captions drawn from the MSCOCO~\cite{lin2014microsoft} training split, and from the CC3M~\cite{cc3m} and CC12M~\cite{cc12m} large scale datasets.

Once the datastore is encoded, given an embedding ($e^I_n$ or $e^T_n$, during inference and training, respectively), the $K$ most similar captions (during inference) or $K+1$ captions (during training) are retrieved. During training, the caption with the highest similarity score is selected as the target for the language model decoder, encouraging the model to learn that similar embeddings should generate the same caption, thereby generalizing from the input image. The remaining $K$ retrieved captions are used to construct the prompt. In our experiments, following SmallCap and LMCap, we set $K=4$ for most evaluations. These captions are then combined into a prompt, as described in the next subsection.

\subsection{Building the Prompt}

To condition caption generation on the retrieved $K$ most similar captions, the following prompt template was used: 
\vspace{-0.5em}
\[
\boxed{
\begin{minipage}{0.9\linewidth}
\tt
Similar images have the following captions: \{caption\textsubscript{1}\} ... \{caption\textsubscript{k}\}.\\[6pt]
Write a caption for this image:
\end{minipage}
}
\]

\begin{figure}
  \centering
  \includegraphics[width=\linewidth]{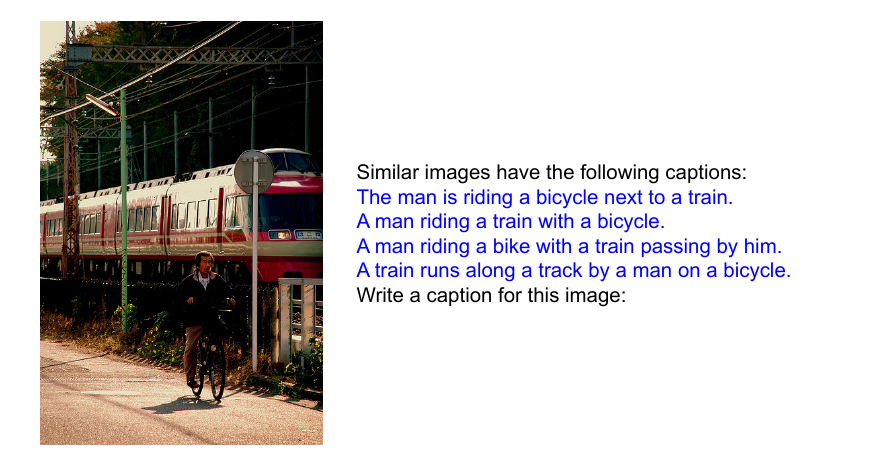}
  \caption{Example illustrating the use of the considered prompt template with a set of retrieved captions (blue).}
  \label{fig:caption_example}
\end{figure}

An example illustrating the use of the previous prompt template is shown in Figure~\ref{fig:caption_example}.
    
The prompt template explicitly provides the model with examples of relevant captions, helping it capture both stylistic and semantic patterns present in similar images. In the majority of our experiments, the order of the captions was fixed and based on decreasing CLIP similarity, but we have also investigated how varying this order may affect result quality. These results are reported in Apendix~\ref{order}.

\subsection{Training the Language Model}

Most of our experiments were conducted using GPT2-base as the decoder, although we also tested the use of larger language model variants. We employed a multi-head cross-attention mechanism to connect the encoder and the decoder, where each layer attends to vectors resulting from the corrected CLIP encodings (i.e., the input vector plus the vectors for the $K$ retrieved captions). Similar to what was done in SmallCap~\cite{ramos2023smallcap}, we adjust the dimensionality of the projection matrices in the cross-attention layers, providing some control over the number of trainable parameters. In our experiments with GPT2-base, we used $h=12$ cross-attention heads, each with a default dimensionality of $d=64$, which we scaled down by a factor of 4.

The textual prompt that was previously described is tokenized and provided as input to the decoder. The decoder then generates a caption conditioned on the CLIP embeddings, and on the retrieved captions from the datastore. During inference, the decoding process employs deterministic beam search with a beam size of five~\cite{cohen2019empirical}.

During training, only the weights of the cross-attention layers, and the Low-Rank Adaptation (LoRA) layers added to the attention projections (keys, values, and queries) of each attention block in the decoder, are updated. Training is performed by optimizing the cross-entropy loss to predict the correct tokens in the most similar retrieved caption, effectively applying teacher forcing. For LoRA, we used the Rank-stabilized LoRA (rsLoRA) variant~\cite{rslora} with a rank of 32 and a $\alpha$ of 32. The optimizer is AdamW with an initial learning rate of $1\text{e-}4$ and a batch size of 32. Training runs for up to 10 epochs with early stopping based on the last three evaluation scores derived from a validation split of the original MSCOCO training split containing 5\% of the data, the scores of which are computed every 2048 steps. Training the decoder requires up to 6 hours on a single NVIDIA RTX 6000 GPU, and consumes up to 32 GB of memory.

A key aspect of our approach relates to using the closest caption from the datastore as the training target, instead of the ground-truth caption. This strategy improves generalization, as previously observed in CLOSE~\cite{gu2022can}, and encourages the generation of more general captions that capture the most important components of the images. Examples of such captions are provided in Section~\ref{sec:appendi_examples}.

%% file: sec/4_evaluation.tex
\section{Experimental Evaluation}
\label{sec:evaluation}

The following section describes the experimental evaluation of the proposed approach.

\subsection{Datasets and Metrics}

To evaluate the proposed approach, we conducted extensive experiments on two well-known English datasets, namely MSCOCO~\cite{lin2014microsoft} and NoCaps~\cite{agrawal2019nocaps}. 

MSCOCO is a widely used benchmark for image captioning, and is therefore ideal for our task. We used the Karpathy~\cite{ioffe2015} splits, which consist of 113k/5k/5k images for training, validation, and testing, respectively. Each image has five human-annotated captions used as the ground truth for evaluation. Since our approach is text-only, we trained our models using only the textual captions from the training split, resulting in approximately 550k training examples.
The second dataset, i.e. NoCaps, contains 15k images spanning nearly 400 classes in novel domains. These images are split into validation and test sets containing 4,500 and 10,600 images, respectively. Generated captions are evaluated by comparing them to 10 human-authored captions per image. Due to the unavailability of the NoCaps online benchmark at the time of writting, we report results only on the corresponding validation set.

The datastore used in our models is similar to that of SmallCap, and it contains captions from the MSCOCO~\cite{lin2014microsoft} training split, from CC3M~\cite{cc3m}, and from CC12M~\cite{cc12m}, totaling approximately 16 million captions. In addition, we used images from Flickr30k~\cite{flikr30k} to compute $\mu_d^{I}$ and $\sigma_d^{I}$, while $\mu_d^{T}$ and $\sigma_d^{T}$ were computed from the datastore itself. This setup allows the model to remain independent of any paired image-caption data.

For model evaluation, we report standard metrics including BLEU-1 (B@1), BLEU-4 (B@4), METEOR (M), and CIDEr (C), which were computed using the official MSCOCO evaluation package.

\subsection{Captioning Results}

\begin{table*}[t!]
\centering
\scriptsize
\begin{tabular}{l c c c c c c c c c c c}
                                               &          &          & \multicolumn{4}{c}{MSCOCO}          & & \multicolumn{4}{c}{NoCaps (CIDEr)} \\
\cline{4-7} \cline{9-12}
Method                                         & Encoder  & Decoder     & BLEU@1    & BLEU@4    & METEOR     & CIDER      & & In    & Near  & Out   & Overall\\
\hline
\multicolumn{12}{l}{\textbf{Training-free approaches}} \\
\hline
ConZic~\cite{zeng2023conzic}                   & ViT-B/32 & BERT-base   &   --   &    1.3 &   11.2 &  13.3 & &    -- &   --  &   --  &   --   \\
ZeroCap~\cite{tewel2022zerocap}                & ViT-B/32 & GPT2-medium &   49.8 &    7.0 &   15.4 &  34.5 & &    -- &   --  &   --  &   --   \\
Socratic~\cite{zeng2022socratic}               & ViT-L/14 & GPT-3       &   --   &    6.9 &   15.0 &  44.5 & &    -- &   --  &  --   &  --    \\
MeaCap\textsubscript{TF}~\cite{zeng2024mea}    & ViT-B/32 & CBART    &   --   &    9.1 &   20.6 &  56.9 & &  35.3 &  39.0 &  45.1 &  40.2  \\
LMCap~\cite{ramos2023lmcap}                    & ViT-H-14 & XGLM-2.9B   &   --   &   19.9 &   22.0 &  75.9 & &    -- &   --  &  --   &   --   \\
\hline
\multicolumn{12}{l}{\textbf{Text-only approaches}} \\
\hline
MAGIC~\cite{su2022language}                    & ViT-B/32 & GPT2-small  &   56.8 &   12.9 &   17.4 &  49.3 & &    -- &   --  &   --  &   --   \\
DeCap~\cite{li2023decap}                       & ViT-B/32 & Transformer &   --   &    8.9 &   17.5 &  50.6 & &  41.9 &  41.7 &  46.2 &  42.7  \\
CLM~\cite{wang2022zero}                        & --       & GPT2        &   59.3 &   15.0 &   18.7 &  55.7 & &    -- &   --  &   --  &   --   \\
MacCap~\cite{qiu2024mining}                    & ViT-B/32 & OPT-1.3B    &   61.4 &   17.4 &   22.3 &  69.7 & &    -- &   --  &   --  &   --   \\
WS-ClipCap~\cite{clipcap}                & --       & GPT2        &   65.5 &   22.1 &   22.2 &  74.6 & &    -- &   --  &   --  &   --   \\
MeaCap\textsubscript{ToT}~\cite{zeng2024mea}   & ViT-B/32 & CBART   &   --   &   17.7 &   24.3 &  84.8 & &  38.5 &  43.6 &  50.0 &  45.1  \\
CapDec~\cite{nukrai2022text}                   & ResNet50 & GPT2-large  &   69.2 &   26.4 &   25.1 &  91.8 & &  60.1 &  50.2 &  28.7 &  45.9  \\
CapDec+RLCF-S~\cite{zhao2023test}              & ViT-B/16 & OPT-125M    &   --   &   --   &   --   &  --   & &  68.3 &  58.5 &  35.3 &   --   \\
ViECap~\cite{fei2023transferable}              & ViT-B/32 & GPT2-base   &   --   &   27.2 &   24.8 &  92.9 & &  61.1 &  64.3 &  65.0 &  66.2  \\
EntroCap~\cite{yan4737282entrocap}             & ViT-B/32 & GPT2-base   &   --   &   27.6 &   25.3 &  94.3 & &  62.5 &  64.5 &  67.5 &  67.0  \\
ViECap+ToCa~\cite{zhou2024text}                & ViT-B/32 & GPT2-base   &   --   &   27.1 &   25.4 &  95.0 & &  64.6 &  69.1 &  70.5 &  70.9  \\
MeaCap\textsubscript{InvLM}~\cite{zeng2024mea} & ViT-B/32 & GPT2-base & -- &   27.2 &   25.3 &  95.4 & &    -- &   --  &   --  &   --   \\
ICSD~\cite{ma2024image}                        & ViT-B/32 & BERT-base   &   --   &   29.9 &   25.4 &  96.6 & &  42.9 &  44.3 &  35.6 &  42.7  \\
CLOSE~\cite{gu2022can}                         & ViT-L/14 & T5-base     &   --   &   29.5 &   25.7 &  97.8 & &    -- &   --  &   --  &   --   \\
ArcSin~\cite{liu2024arcsin}                    & ViT-L/14 & T5-base     &   --   &   30.3 &   --   &  99.6 & &    -- &   --  &   --  &   --   \\
SynTIC~\cite{liu2024improving}                 & ViT-B/32 & Transformer &   --   &   29.9 &   25.8 & 101.1 & &    -- &   --  &   --  &   --   \\
TipCap~\cite{wang2024text}                     & ViT-L/14 & GPT2-large  &   73.3 &   \textbf{31.4} &   \textbf{26.9} & 106.6 & &  \textbf{80.2} &  62.3 &  39.6 &  60.3  \\
IFCap~\cite{ifcap2024}                             & ViT-B/32 & GPT2-base   &   --   &   30.8 &   20.3 & 108.0 & &  70.1 &  72.5 &  72.1 &  74.0 \\
\hline
TOMCap (no training, retrieval)               & ViT-L/16 & GPT2-base    &   30.4   &   8.6   &   8.4   &  15.2   & &  --&   --  &   --  &   --   \\
TOMCap (retrieval only)                & ViT-L/16 & GPT2-base   &   72.4   &   28.2   &   25.2   &  101.6   & &    -- &   --  &   --  &   --   \\
TOMCap (embedding only)                       & ViT-L/16 & GPT2-base   &   64-1   &   19.1   &   22.1   &  76.6   & &    -- &   --  &   --  &   --   \\
TOMCap              & ViT-L/16 & GPT2-base   &   72.7   &   28.4   &   25.8   &  103.4   & &    76.3 &   75.5  &   75.7  &   76.2   \\
TOMCap (large)                                        & ViT-L/16 & GPT2-large   &   \textbf{73.9}   &   30.2   &   26.6   &  \textbf{108.3}   & &    77.1 &   \textbf{76.6}  &   \textbf{75.9}  &   \textbf{76.4 }  \\

\hline
\end{tabular}
\caption{Results of multiple models on the MSCOCO and NoCaps datasets.}
\label{tab:results}
\end{table*}

Table~\ref{tab:results} provides the main experimental results obtained when evaluating our approach on the MSCOCO and NoCaps datasets. The presented results make use of the best hyperparameters obtained from ablation studies on the model. We also compare the results obtained with previous training-free and text-only training models, retrieved from the corresponding publications, as well as with ablated versions of TOMCap, as detailed next:

\begin{itemize}
\item TOMCap (no training, retrieval): A version similar to the original LMCap method~\cite{ramos2023lmcap} (without CLIP re-ranking), where no training is required. As such, the model is limited to using the prompt to guide generation

\item TOMCap (retrieval only): A trained version that does not contain the cross-attention mechanism, where the generation process is only conditioned on the $K$ most similar captions that compose the prompt.

\item TOMCap (embedding only): A version that makes use of the cross-attention mechanism, where the prompt consists of a simple template that does not use any similar captions to guide the generation.

\item TOMCap: A version with the full model implementation, applying all modality gap reduction techniques.

\item TOMCap (large): A version that improves upon the base TOMCap model (which employs GPT2-base as the decoder, as all the previously mentioned versions) by utilizing a larger decoder, namely GPT2-large.

\end{itemize}

The obtained experimental results show that TOMCap outperforms all previous zero-shot methods and most text-only approaches, although still achieving scores that significantly lag behind those of specialized models that require image-caption pairs for supervised training.

When compared to other text-only training approaches, TOMCap outperforms methods such as MAGIC~\cite{su2022language}, DeCap~\cite{li2023decap}, and CLOSE~\cite{gu2022can} in the MSCOCO dataset. On the NoCaps dataset, our model performs on par with or surpasses other approaches in terms of the ability to generate captions in domains that are not present in the training set. This effect is particularly evident in the out-of-domain subset of NoCaps, where retrieval compensates for the lack of domain-specific training data.

The results obtained from comparing a version of our model that uses only embeddings, with the version that also relies on the $K$ retrieved captions, reinforces previous findings that employing a separate datastore to retrieve semantically similar captions is an effective strategy to improve image captioning performance~\cite{ramos2023smallcap}. 

\subsection{Impact of Retrieval Augmentation}
\label{sec:retr}

Besides assessing captioning quality, we also examined the impact of different model configurations, focusing specifically on the retrieval augmentation step, which is the most impactful in terms of result quality. Our model allows for changes in the configuration of the number of captions retrieved ($K$), and on the standard deviation of the noise injected during the retrieval process ($L$), the latter being used to further reduce the modality gap.

Regarding the number of captions retrieved per image ($K$), we tested values of 4, 6, and 8. Additionally, for $K=6$ and $K=8$, we tested a variant where the model was trained using fewer retrieved captions, only increasing the value of $K$ during inference. This was done to test whether allowing a broader range of captions during inference, when compared to training, could significantly affect our model. Results on the MSCOCO dataset for the GPT2-base decoder are reported in Table~\ref{tab:topk}.

\begin{table}[t!]
\centering
\footnotesize
\begin{tabular}{l c c c c}
Retrieved Captions                      & B@1  & B@4  &  METEOR   & CIDER     \\
\hline
$K=4$                                   & 72.7 & 28.4 & 25.8 & 103.4   \\
$K=6$                                   & 73.5 & 28.9 & 25.9 & 104.0   \\
$K=6$ (train $K=4$)                   & 73.4 & 29.4 & 26.2 & 105.6   \\
$K=8$                                   & 73.0 & 28.9 & 25.9 & 104.2   \\
$K=8$ (train $K=6$)                   & 74.1 & 29.7 & 26.1 & 105.4   \\
\hline
\end{tabular}
\caption{Results on MSCOCO when varying the number of retrieved captions $K$, with GPT2-base as the decoder.}
\label{tab:topk}
\end{table}

The results from the tests show that there is no significant impact on increasing the number of captions retrieved during training. It is possible to observe that despite doubling the number of captions retrieved from 4 to 8, this only results in an improvement of 0.8 in terms of the CIDEr score. Furthermore, we show that training the model with a lower value of $K$ (e.g., $K=4)$ and then increasing this value actually yields better results. For the remaining tests, we have opted to fix the value of $K$ at 4.

A second set of experiments focused on the magnitude of the Gaussian noise injected into the model during the retrieval step (L). To ensure comparability, the number of retrieved captions was fixed at $K=4$ and the scaling factor $B$ was set at $B=0.125$. The noise levels that were tested correspond to $L \in \{0, 0.1, 0.125, 0.5, 1\}$, where $L=0$ corresponds to using no noise injection during training. The injected noise $n$ is sampled from a Gaussian distribution scaled by $L$, i.e., $n \sim \mathbf{N}(0,L)$. This noise is applied after the correction operation, meaning that the modality gap has already been minimized. The results from these experiments are reported in Table~\ref{tab:l_val}.

Setting $L=0.1$ yields the best performance among the values that were tested. Notably, the performance degrades significantly as the magnitude of the injected noise increases. Despite $L=0.1$ achieving the highest CIDEr scores on MSCOCO, the absence of noise ($L=0$) also produces comparable results, suggesting that the residual modality gap after correction is not significant. Furthermore, these experiments demonstrate that even under conditions with relatively high noise injection, the model retains a reasonable performance, highlighting its robustness to noise in the cross-attention layers of the decoder.

To further assess the importance of the retrieval step, we conducted experiments to evaluate the impact of the similarity and diversity of the retrieved captions, on the quality of the generated captions, the results from these experiments are reported on Apendix ~\ref{sec:mmr}.

\begin{table}[t!]
\centering
\footnotesize
\begin{tabular}{l c c c c}
Variance                     & BLEU@1  & BLEU@4  &  METEOR   & CIDER     \\
\hline
$L=0$ (no noise)                        & 72.9 & 28.7 & 25.8 & 103.0   \\
$L=0.1$                                 & 72.7 & 28.4 & 25.8 & 103.4   \\
$L=0.125$                               & 72.0 & 27.7 & 25.8 & 101.5   \\
$L=0.5$                                 & 69.3 & 25.0 & 24.9 & 93.7   \\
$L=1$                                   & 69.1 & 25.1 & 24.7 & 93.4   \\
\hline
\end{tabular}
\caption{Results on MSCOCO when varying the variance of the injected noise into the retrieval step.}
\label{tab:l_val}
\end{table}

\subsection{Impact of the Modality Gap Reduction}

In addition to studying the impact of retrieval, it is important to evaluate the effects of the modality gap reduction measures, namely the mean correction and the standard deviation rescaling. To this end, we tested our model using variants of the correction function, applying either just one or none of these methods. Table~\ref{tab:gap_red_measures} presents three versions of our model, all trained with $B=0.125$, $L=0.1$ and $K=4$, whilst applying the mean-var correction. The first version does not perform any correction. The second version, similarly to GR-CLIP~\cite{MixedModalitySearch}, applies only the mean correction to align the embeddings of images and texts. Finally, the third version incorporates both the mean and the standard deviation corrections.

The results demonstrate that the modality gap reduction mechanism is crucial to the model’s success. We attribute the improvement primarily to the retrieval process, which becomes capable of retrieving captions that are more semantically aligned after the correction. This effect propagates throughout the model, as the improved retrieval quality influences the selection of training targets and, consequently, enhances overall performance.

\begin{table}[t!]
\centering
\footnotesize
\begin{tabular}{l c c c c}
Correction                      & B@1  & B@4  &  METEOR   & CIDER     \\
\hline
No correction                           & 67.1 & 19.8 & 22.4 & 79.3   \\
Only mean corrected                     & 72.0 & 27.7 & 24.8 & 102.2   \\
Std and mean corrected                  & 72.7 & 28.4 & 25.8 & 103.4 \\
\hline
\end{tabular}
\caption{Results on MSCOCO when changing the corrections applied to address the modality gap.}
\label{tab:gap_red_measures}
\end{table}

Another ablation study focused on the impact of the magnitude of the noise that is injected into the embeddings used by the decoder. It is not immediately clear whether the decoder and the retrieval system should receive the same magnitude of noise. Following a similar methodology to the analysis that was made in Section~\ref{sec:retr} for $L$, we tested different values for $B$ while keeping $L=1$, $K=4$, and applying all proposed modality gap corrections. The decoder used in these experiments was GPT2-base. Specifically, we evaluated $B\in\{0,0.1,0.125,0.5,1\}$, with results being reported in Table~\ref{tab:B_val}.

The results indicate that the noise injected into the decoder embeddings has a minor impact on performance. This is evidenced by the small difference of only 1.5 points in CIDEr score between the worst-case scenario (no noise injected) and the best-performing magnitude ($B=0.125$), suggesting that the decoder is very robust to variations in this noise component.

\begin{table}[t!]
\centering
\footnotesize
\begin{tabular}{l c c c c}
Variance                      & BLEU@1  & BLEU@4  &  METEOR   & CIDER     \\
\hline
$B=0$ (no noise)                        & 72.4 & 28.2 & 25.7 & 101.9   \\
$B=0.1$                                 & 72.5 & 28.3 & 25.7 & 102.9   \\
$B=0.125$                               & 72.7 & 28.4 & 25.8 & 103.4   \\
$B=0.5$                                 & 72.6 & 28.5 & 25.7 & 103.2   \\
$B=1$                                   & 72.7 & 28.6 & 25.8 & 103.0   \\
\hline
\end{tabular}
\caption{Results on MSCOCO when varying the variance of the injected noise into the GPT2 decoder.}
\label{tab:B_val}
\end{table}

\begin{figure*}[t!]
  \centering
  \includegraphics[width=1.0\textwidth]{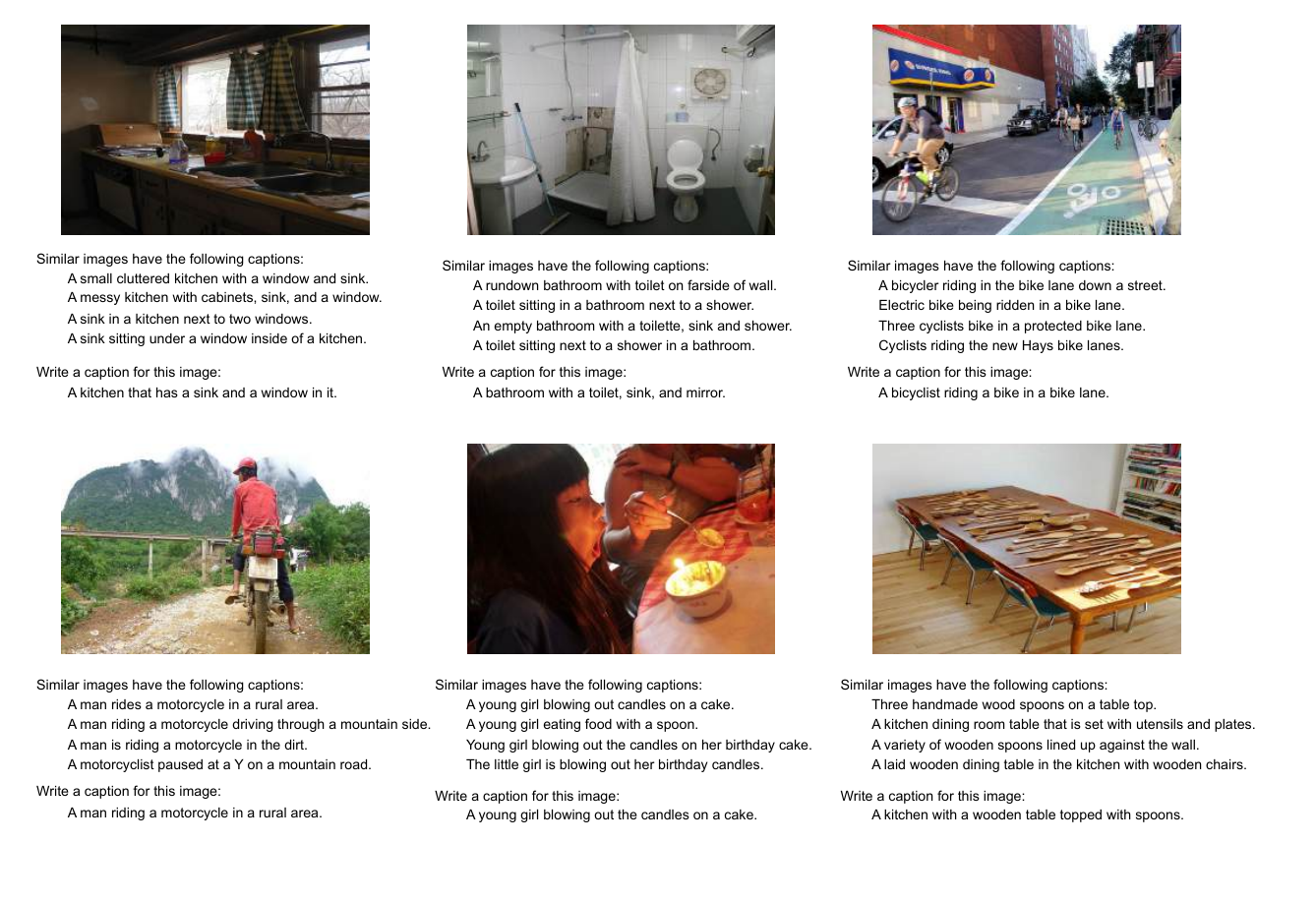}
  \caption{Examples of captions retrieved and generated by TOMCap.}
  \vspace{-0.4cm}
  \label{fig:examples}
\end{figure*}

\subsection{Qualitative Examples}
\label{sec:appendi_examples}

Figure \ref{fig:examples} presents some examples of captions obtained by the retrieval process and generated by the decoder. The images in question were taken from the validation split of the MSCOCO dataset. For the generation of captions for these images, the model utilized the GPT2-base decoder and the following configuration: $K=4$, $B=0.125$, and $L=0.1$.

%% file: sec/5_conclusion.tex
\section{Conclusions and Future Work}
\label{sec:conclusion}

This paper presented TOMCap, i.e. a text-only training method for image captioning that leverages a pre-trained GPT-2 language model as a decoder, which is prompted with retrieved captions, together with a SIGLIP2 model as an encoder. The encoder is used together with a modality gap correction mechanism that approximates the means and standard deviations of image and text embeddings.

Experimental results demonstrate that TOMCap is highly competitive with state-of-the-art text-only methods and robust in cross-domain settings. Nevertheless, there are some opportunities for further improvements. A major limitation lies in the assumption that the modality gap operates independently across embedding dimensions. Developing approaches that relax this assumption could further reduce the modality gap and improve performance. It is also important to note that our experiments were conducted solely on English-language captions. Future work will explore training TOMCap in other languages. We hypothesize that TOMCap is a promising approach when employed for less-resourced languages because our approach enables training without requiring image-text pairs. This aligns with one of the primary motivations for text-only captioning systems, in enabling multilingual captioning in scenarios where large-scale paired datasets are not available.

%% file: sec/appendix.tex
\appendix

\section{Limitations and Ethical Considerations}

The work presented in this paper does not raise major ethical concerns, as all experiments were conducted on well-known public datasets specifically designed for evaluating image captioning methods. However, there are important limitations to consider.

We for instance have that the evaluation was conducted exclusively in English. As a result, performance on other languages was not assessed. While our model can be adapted for multilingual captioning, it would require the use and fine-tuning of a different decoder, and also the use of multilingual data. One potential solution would be to use a language model to translate existing English datasets, such as MSCOCO or the Conceptual Captions datasets, into other languages. Since TOMCap does not require paired image-caption data, improvements in multilingual decoders can directly address this limitation.

Another concern is the potential for biases in the datasets being used, particularly from widely used sources such as MSCOCO. Although our model is trained using MSCOCO, the target captions are drawn from a larger datastore that includes captions from multiple datasets, which can help mitigate dataset-specific biases. Expanding the datastore further could help to reduce biases without the need for exact image-caption pairs. Additionally, biases can also arise from the pretrained CLIP and GPT2 models. While our modular approach allows for easy updates to the encoder and decoder, we recommend careful evaluation of these pretrained components before deploying the model.

\section{Modality Gap Analysis}

\begin{figure*}[t!]
    \centering
    \begin{minipage}[b]{0.30\textwidth}
        \centering
        \includegraphics[width=\textwidth]{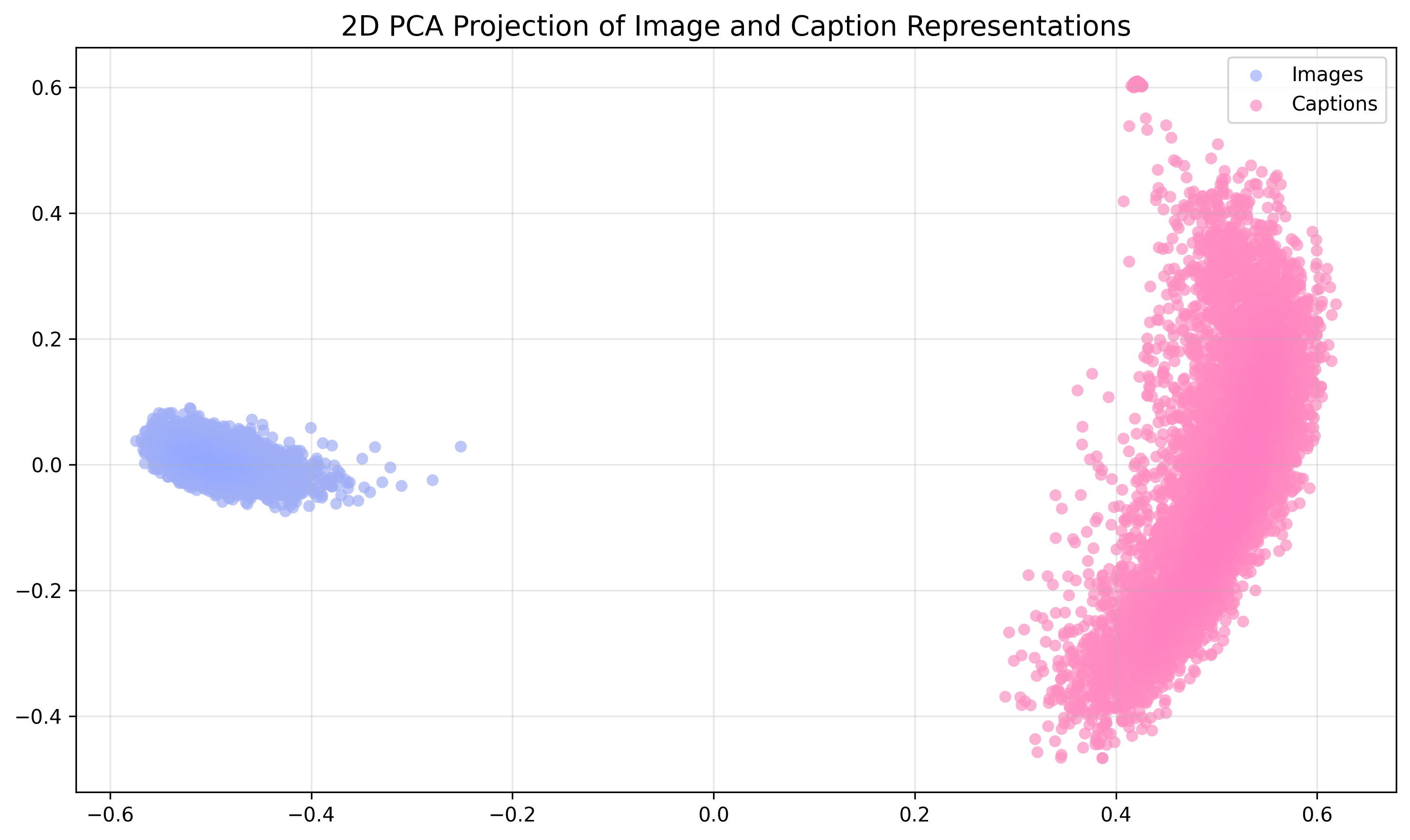}
        \caption*{No correction.}
    \end{minipage}
    \begin{minipage}[b]{0.30\textwidth}
        \centering
        \includegraphics[width=\textwidth]{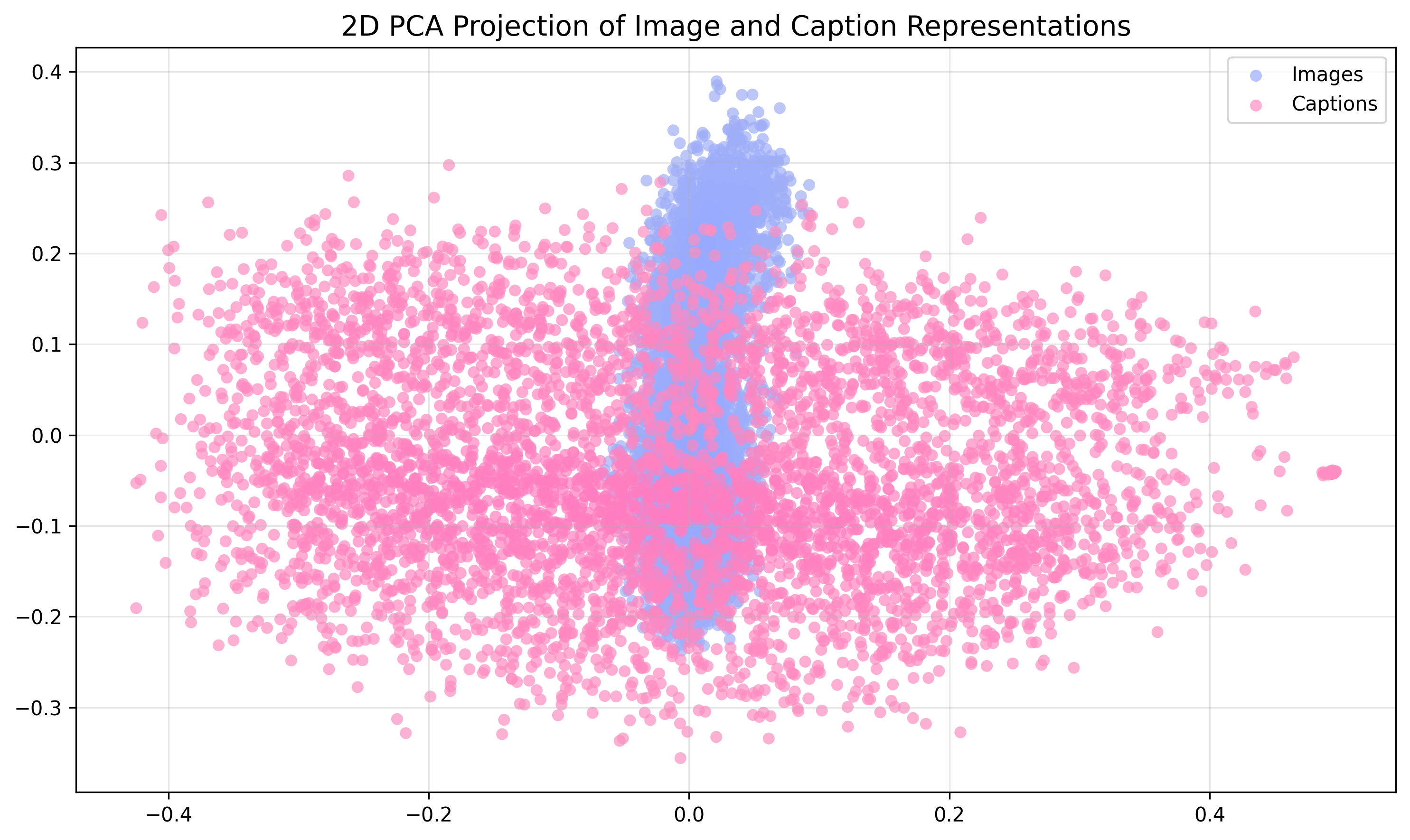}
        \caption*{Only mean correction.}
    \end{minipage}
    \begin{minipage}[b]{0.30\textwidth}
        \centering
        \includegraphics[width=\textwidth]{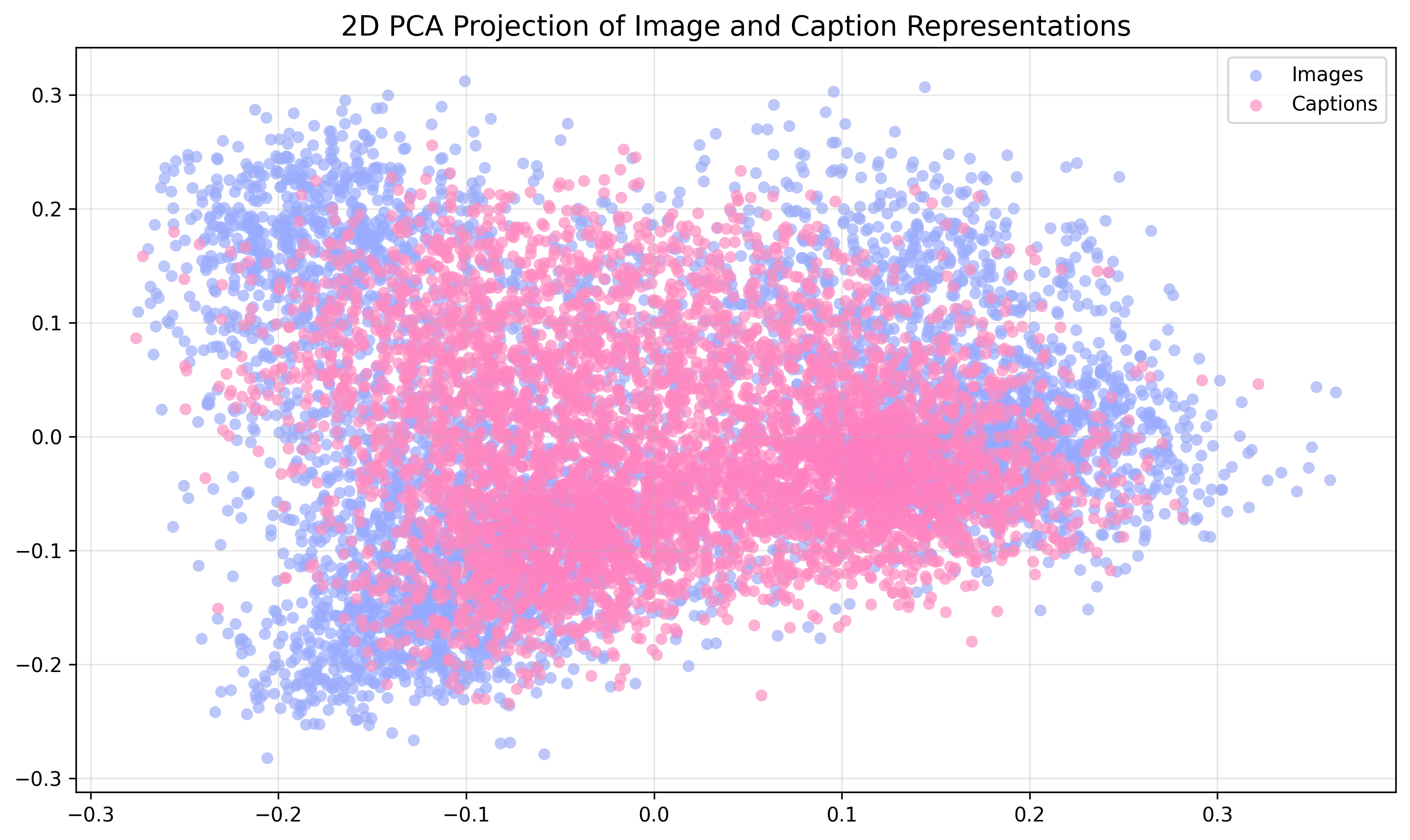}
        \caption*{Mean and standard deviation.}
    \end{minipage}
    \caption{Three t-SNE plots corresponding to different strategies for handling the modality gap.}
    \label{fig:t-sne}
\end{figure*}

\begin{figure*}[t!]
    \centering
    \begin{minipage}[b]{0.49\textwidth}
        \centering
        \includegraphics[width=\textwidth]{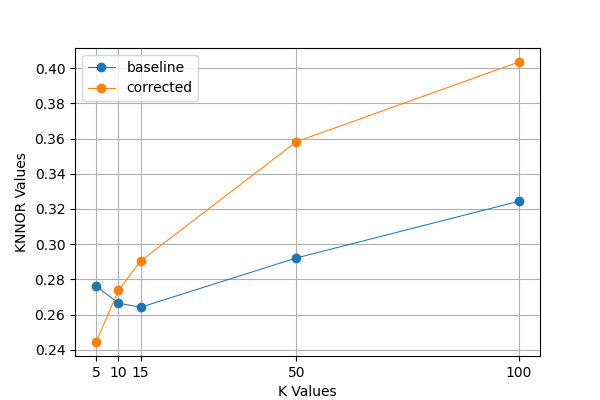}
    \end{minipage}
    \begin{minipage}[b]{0.49\textwidth}
        \centering
        \includegraphics[width=\textwidth]{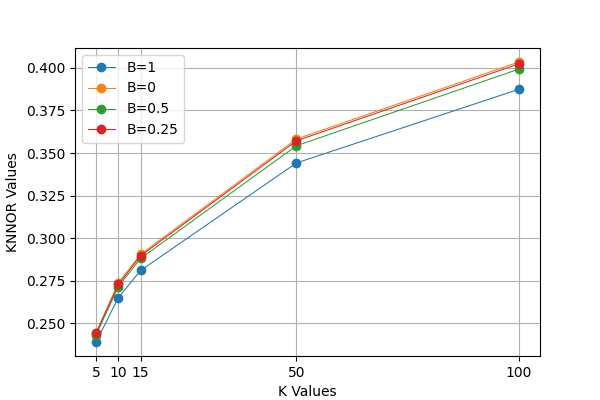}
    \end{minipage}
    \caption{KNOR scores for $k\in\{5,10,15,50,100\}$. On the left, we use a comparison between not applying any correction ("baseline") and applying the mean/standard deviation correction. On the right, we show the impact of varying the Gaussian noise magnitude.}
    \label{fig:knor}
\end{figure*}

Due to the impact of the modality gap on the performance of text-only training approaches, a detailed analysis of the modality gap, and of the methods used to reduce it, is necessary. Our experimental results indicate that the modality gap has a significant effect on captioning quality. To address this, we conducted experiments aimed at creating a training environment that closely resembles the conditions encountered during inference. All experiments reported here were conducted using the MSCOCO validation dataset.

We began by analyzing t-SNE projections of the embeddings as can be seen in Figure~\ref{fig:t-sne}. These projections reveal that the initial CLIP embeddings obtained directly from SigLIP2~\cite{tschannen2025siglip2multilingualvisionlanguage}, seen on the left of Figure~\ref{fig:t-sne}, exhibit a substantial modality gap that must be mitigated to improve performance. The middle plot in Figure~\ref{fig:t-sne} indicates that using the means of different text and image instances, as explored in previous work~\cite{MixedModalitySearch}, effectively reduces the modality gap. However, a scaling mismatch remains: although the means of the two distributions are close, their overall shapes differ. To address this, we applied a rescaling based on the standard deviations in both datasets (i.e., over the images and the texts). This approach further reduces the modality gap, as shown in the right plot of Figure~\ref{fig:t-sne}. This is achieved by leveraging the fact that, given two Gaussian distributions with known means and standard deviations, a value from one distribution can be transformed into a corresponding value of the other. The main limitation of this approach is the assumption that the CLIP embedding vectors follow independent Gaussian distributions, which does not fully hold in practice.

To further investigate the modality gap, we evaluated the $k$-Nearest Neighbors Overlap Ratio (KNOR)~\cite{knor}, which measures how well the local geometry and semantic relationships in the embeddings are preserved. For this analysis, we considered the 2,500 captions comprising the MSCOCO validation dataset and tested values of $k\in\{5, 10, 15, 50, 100\}$. This allows us to examine the overlap between the $k$ captions retrieved during inference (i.e., using image embeddings) and those retrieved during training (i.e., using textual embeddings).

The procedure involves two retrieval steps, namely one using image embeddings as keys, and another using textual embeddings as keys. The intersection of the two sets of retrieved captions is computed and divided by $k$, and this calculation is repeated for all 2,500 captions in the dataset. Results for different values of $k$ are shown in Figure~\ref{fig:knor}, specifically on the left plot.

The results indicate that our correction methods perform similarly for smaller values of $k$ but, as $k$ increases, the overlap achieved by our approach surpasses that of using no correction. This suggests that, while the closest captions may not exactly match the image, they are still positioned near the image embedding in the corrected space.

Additionally, noise injections appear to have a minimal impact on retrieval performance, even though they simulate the modality gap during the training procedure, as shown in the right-side plot of Figure~\ref{fig:knor}.

\section{Positional Biases in Retrieved Captions}
\label{order}

Our model employs a language model decoder based on GPT-2, which is autoregressive and position-aware through the use of learned absolute positional embeddings. As a result, the order in which retrieved captions are presented in the prompt may introduce positional biases during training or inference. To evaluate this potential issue, we conducted experiments in which the order of captions within the prompt was systematically varied. The results of these experiments are summarized in Table~\ref{tab:order}.

Empirical results indicate that the ordering of the $K$ retrieved captions has a negligible impact on the model’s ability. More specifically, the model was trained using captions arranged in descending order of similarity between the input embedding and the retrieved embeddings, while during inference the caption order was altered according to different configurations, as detailed in Table~\ref{tab:order}. The observed consistency across configurations suggests that GPT-2’s positional embeddings do not introduce significant bias in this retrieval-augmented generation setup.

\begin{table}[t!]
\centering
\footnotesize
\begin{tabular}{l c c c c}
Retrieved Captions                      & BLEU@1  & BLEU@4  &  METEOR   & CIDER     \\
\hline
Decreasing                              & 72.7 & 28.4 & 25.8 & 103.4   \\
Increasing                              & 73.0 & 28.7 & 25.8 & 103.2   \\
Random                                  & 72.4 & 28.8 & 25.8 & 103.9   \\
\hline
\end{tabular}
\vspace{-0.2cm}
\caption{Results on MSCOCO when switching the order of presentation for the retrieved captions in the prompt.}
\label{tab:order}
\vspace{-0.25cm}
\end{table}

\section{Impact of Re-Ranking}
\label{sec:mmr}

\begin{table}[b!]
\centering
\footnotesize
\begin{tabular}{l c c c c}
MMR Settings                      & BLEU@1  & BLEU@4  &  METEOR   & CIDER     \\
\hline
$\lambda = 0.5$                         & 66.8 & 23.9 & 24.2 & 90.3   \\
$\lambda = 0.2$                         & 67.2 & 24.7 & 24.5 & 92.7   \\
$\lambda = 0$ (default)                 & 72.7 & 28.4 & 25.8 & 103.4  \\
$\lambda = -0.2$                        & 68.4 & 24.5 & 24.6 & 92.6   \\
$\lambda = -0.5$                        & 68.5 & 25.0 & 24.6 & 93.5   \\
\hline
\end{tabular}
\caption{Results of applying MMR to the retrieved captions for various values of $\lambda$.}
\label{tab:mmr}
\end{table}

To further assess the importance of the retrieval step, we conducted experiments to evaluate the impact of the similarity and diversity of the retrieved captions, on the quality of the generated captions. To this end, we applied a re-ranking operation to the top 16 retrieved captions, based on the Maximal Marginal Relevance (MMR)~\cite{carbonell1998use} criterion, defined as follows:
\begin{equation}
    \text{SP}(D_i) =  \, \text{sim}(D_i, Q),
\end{equation}
\begin{equation}
    \text{DSP}(D_i) = \lambda \, \max_{D_j \in S} \text{sim}(D_i, D_j),
\end{equation}
\begin{equation}
    \text{MMR} = \arg\max_{D_i \in R \setminus S} \big[ \text{SP}(D_i) - \text{DSP}(D_i) \big],
\end{equation}
where $D_i \in R$ denotes a candidate caption from the retrieved set, $D_j \in S$ denotes a caption already selected, $Q$ represents the input CLIP embedding, and $\lambda \in [0,1]$ is a hyperparameter controlling the balance between similarity and diversity among captions in the re-ranked set $S$. The results of the tests conducted to assess the use of MMR are presented in Table~\ref{tab:mmr}.

Overall, we see that adding a re-ranking operation during the retrieval step does not lead to a measurable improvement in performance, either by promoting coherence or diversity in the retrieved set.